\documentclass[10pt,twocolumn,letterpaper]{article}

\usepackage{iccv}              

\usepackage{booktabs}
\usepackage{multirow}
\usepackage{colortbl}
\usepackage{xcolor}
\usepackage{graphicx}
\usepackage{array}
\usepackage{adjustbox}
\usepackage{enumitem}
\usepackage{comment}

\usepackage{tcolorbox}
\definecolor{grey}{rgb}{0.85,0.85,0.85}
\newtcolorbox{mybox}[2]{
    arc=3pt,
    boxrule=#2pt,
    colback=#1,
    width=8.25cm,
    halign=left,
}

\newcommand*{\our}{\texttt{E-FineR}}

\definecolor{iccvblue}{rgb}{0.21,0.49,0.74}
\usepackage[pagebackref,breaklinks,colorlinks,allcolors=iccvblue]{hyperref}

\def\paperID{*****} 
\def\confName{ICCV}
\def\confYear{2025}

\title{
Vocabulary-free Fine-grained Visual Recognition \\
via Enriched Contextually Grounded Vision-Language Model}

\author{Dmitry Demidov \\
{\tt\small dmitry.demidov@mbzuai.ac.ae}
\and
Zaigham Zaheer \\
{\tt\small zaigham.zaheer@mbzuai.ac.ae}
\and
Omkar Thawakar \\
{\tt\small omkar.thawakar@mbzuai.ac.ae}
\and
Salman Khan \\
{\tt\small salman.khan@mbzuai.ac.ae}
\and
Fahad Shahbaz Khan \\
{\tt\small fahad.khan@mbzuai.ac.ae}
\\
\and
Mohamed bin Zayed University of Artificial Intelligence \\
UAE, Abu Dhabi
}

\begin{document}
\maketitle
\begin{abstract}

    Fine-grained image classification, the task of distinguishing between visually similar subcategories within a broader category (e.g., bird species, car models, flower types), is a challenging computer vision problem.
    Traditional approaches rely heavily on fixed vocabularies and closed-set classification paradigms, limiting their scalability and adaptability in real-world settings where novel classes frequently emerge. 
    Recent research has demonstrated that combining large language models (LLMs) with vision-language models (VLMs) makes open-set recognition possible without the need for predefined class labels.
    However, the existing methods are often limited in harnessing the power of LLMs at the classification phase, and also rely heavily on the guessed class names provided by an LLM without thorough analysis and refinement. 
    %
    %
    To address these bottlenecks, we propose our training-free method, Enriched-FineR (or \our{} for short), which demonstrates state-of-the-art results in fine-grained visual recognition while also offering greater interpretability, highlighting its strong potential in real-world scenarios and new domains where expert annotations are difficult to obtain.
    Additionally, we demonstrate the application of our proposed approach to zero-shot and few-shot classification, where it 
    demonstrated performance on par with the existing SOTA while being training-free and not requiring human interventions.
    Overall, our vocabulary-free framework supports the shift in image classification from rigid label prediction to flexible, language-driven understanding, enabling scalable and generalizable systems for real-world applications.
    Well-documented code 
    is available on \href{https://github.com/demidovd98/e-finer}{https://github.com/demidovd98/e-finer}.

\end{abstract}    
\begin{figure}[!h]
  \centering
  \includegraphics[width=0.80\columnwidth]{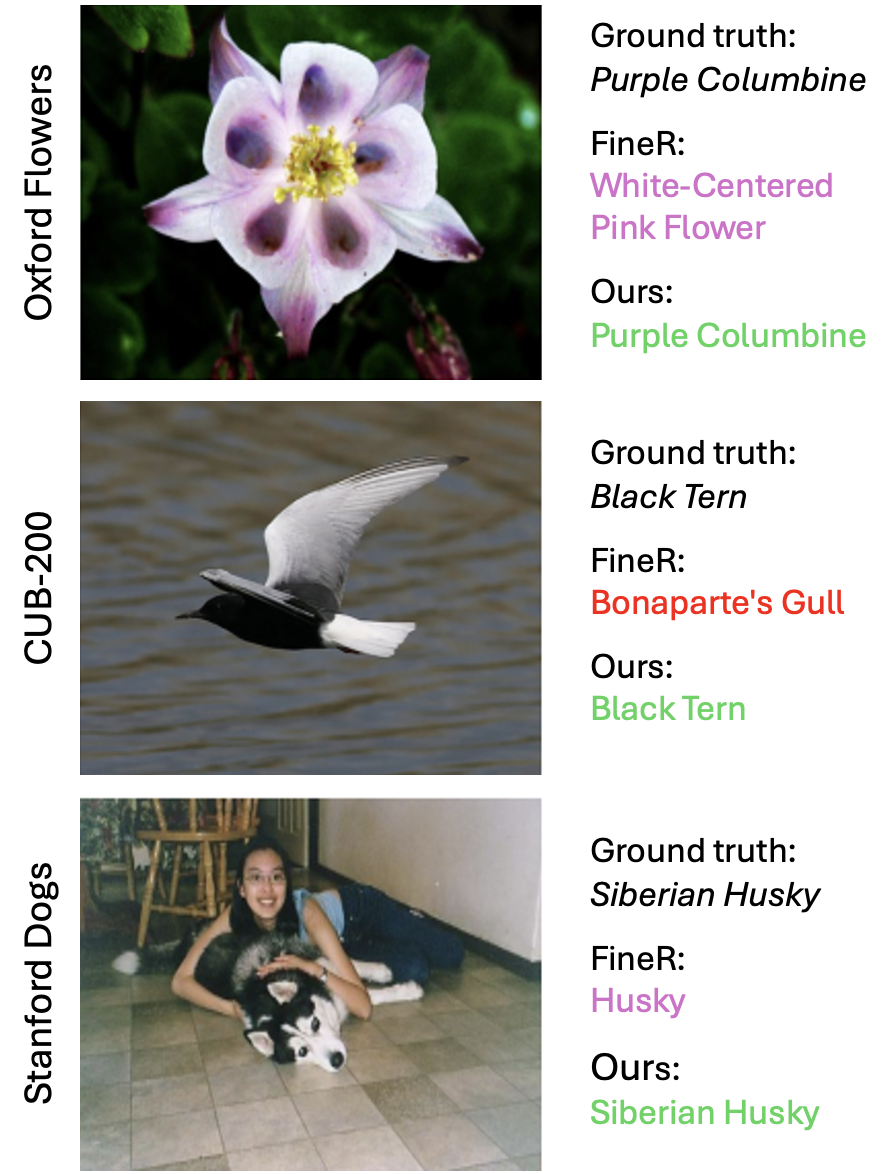}
  \vspace{-7.0pt}
  \caption{Qualitative comparison between the previous state-of-the-art method and our approach on CUB-200, Stanford Cars, and Oxford Flowers datasets. 
  Predictions in green, pink, and purple are correct, partially correct, and incorrect, respectively.
  }
  \label{fig:qual_predictions}
  \vspace{-10.0pt}
\end{figure}

\section{Introduction}
\label{sec:intro}

Fine-grained visual recognition (FGVR) focuses on distinguishing between highly similar subordinate categories within a broader class, such as specific species of birds, flowers, or dog breeds. \cite{Wang2021FeatureFV}. 
While vision-language models like CLIP \cite{CLIP} 
have demonstrated strong generalization in zero-shot classification, adapting them to vocabulary-free scenarios, where class names are not predefined and can emerge dynamically, remains an open challenge.

The existing approaches to this task face notable limitations. 
First, class names alone often provide insufficient semantic grounding for disambiguating visually similar categories. In fine-grained settings, many classes differ by only subtle visual cues (e.g., beak shape, petal arrangement, or fur texture in the case of birds). 
Recent works, such as FineR \cite{finer}, attempt to address this problem by discovering class names and associating them with images via text-to-image similarity. 
However, when distinguishing between visually similar subcategories, the visual context may be as important as the main object due to subtle inter-class variations.
Without richer semantic context, vision-language models may struggle to associate images with the correct class, leading to confusion between semantically or visually adjacent categories.
Second, in vocabulary-free setups, training-based approaches become unreliable. Because class labels must be discovered automatically, they are often noisy or incomplete. Methods that rely on learning from such labels, such as few-shot learners \cite{zhou2022conditionalpromptlearningvisionlanguage, khattak2023maplemultimodalpromptlearning} or prompt-tuned models \cite{khattak2023selfregulatingpromptsfoundationalmodel, cupl}, are prone to propagating errors during training. This undermines the very purpose of vocabulary-free recognition, where classes are not assumed to be fixed or curated.
Another limitation is reliance on manual prompt engineering, which presents a scalability bottleneck. In vocabulary-free scenarios, where the class names are not pre-defined, it is infeasible to handcraft prompts for every possible class. This limits the generality of prompt-based zero-shot and few-shot methods \cite{zhou2022large,menon2022visual}, especially when dealing with long-tailed or unseen distributions.
Finally, most existing methods are built around narrow task assumptions, such as purely zero-shot or purely few-shot settings. Adapting these approaches to new tasks often requires retraining, prompt re-engineering, or dataset-specific tuning, none of which scale to real-world deployments where task structure and label space are often unknown in advance.

To address these limitations, we introduce \our{}, a fully automated and training-free framework for vocabulary-free fine-grained recognition. Our approach enhances vision-language alignment through two key components: (1) the automated generation of class-specific contextual descriptions using large language models, 
and (2) an advanced class name filtration mechanism that softly retains multiple semantically plausible labels, instead of enforcing rigid top-1 predictions. These components improve the model’s semantic grounding and robustness to ambiguity, particularly critical for visually similar categories.
Importantly, \our{} preserves the flexibility of CLIP's zero-shot formulation: no retraining is required, manual prompt engineering is eliminated, and new classes can be added or removed dynamically at inference time. This enables seamless application across vocabulary-free, zero-shot, and few-shot tasks within a unified framework.
In addition to state-of-the-art results in the vocabulary-free setting (see Fig.~\ref{fig:qual_predictions} for example cases), we demonstrate that \our{} yields on par performance with best zero-shot and few-shot methods \cite{zhou2022large,menon2022visual} without any training or manual intervention, demonstrating the efficacy of context-aware naming and flexible label selection for vocabulary-free fine-grained recognition.

Summarily, below are the key highlights of our work:
\begin{itemize}[noitemsep,topsep=0pt, wide=10pt, leftmargin=0pt]
    \item[-] \textbf{Class-specific Contextual Grounding.}
    We propose a novel mechanism to automatically generate class-specific, in-context descriptions for each candidate label using a large language model (LLM). These context-rich sentences are used to query the vision-language model (VLM), significantly improving semantic grounding and enabling better discrimination between visually similar fine-grained categories without requiring 
    prompt engineering or supervision.
    \item[-] \textbf{Advanced Class Name Filtration.}
    We introduce a soft, multi-candidate class name filtering strategy that replaces traditional top-1 or threshold-based selection. Our method retains semantically plausible labels that may not have strong visual coupling but carry relevant descriptive information. This design improves classification recall, reduces premature elimination of useful candidates, and increases robustness in vocabulary-free recognition.
    \item[-] \textbf{Refined Vocabulary-free Recognition Pipeline.}
    We present a fully automated, training-free framework that integrates improved data augmentation, refined CLIP-based pseudo-labelling, and a multi-modal fusion strategy. These enhancements enable our system to dynamically construct unsupervised or partially-supervised classifiers at inference time, supporting flexible deployment across vocabulary-free, zero-shot, and few-shot settings without retraining or manual adaptation.
\end{itemize}




\section{Related Work}

\subsection{Fine-grained Visual Recognition}

Fine-grained visual recognition (FGVR) focuses on classifying visually similar subordinate categories within a broader super-category (e.g., bird species or car models). Due to subtle visual differences between classes, traditional FGVR methods rely not only on labelled data but also on expert-provided auxiliary annotations. These methods typically fall into two categories: (i) feature-encoding approaches that enhance recognition through advanced feature extraction techniques (e.g., bilinear pooling), and (ii) localization-based approaches that identify and focus on the most discriminative regions of an image. More recent methods, such as TransHP \cite{TransHP} and V2L \cite{V2L} integrate vision-language modeling and learn soft prompts to enhance fine-grained discrimination. However, all of these approaches assume access to predefined fine-grained labels. 

\subsection{Vocabulary-free Image Classification}

Compared to typical FGVR approaches, vocabulary-free methods do not require expert annotations or fixed label sets. One of such works is CLEVER \cite{DBLP:journals/corr/abs-2111-03651}, that attempts annotation-free FGVR by extracting non-expert image descriptions and matching them to external text sources. 
Another method, FineR \cite{finer}, shares this goal but differs in that it uses large language models (LLMs) to reason about visual descriptions and infer class names directly without training any additional model.




\subsection{LLM-enchaced Visual Recognition}

The rise of foundation models has significantly advanced tasks like image classification and visual question answering (VQA) through the integration of vision and language, often via LLMs or external knowledge bases. Recent classification methods \cite{cupl,menon2022visual,lin2023train} enhance recognition by converting class names into rich textual descriptions using LLMs like GPT-3, and then leveraging CLIP for image classification. Similarly, modern VQA systems \cite{yang2022empiricalstudygpt3fewshot} utilize image captions from pre-trained models and combine them with questions and in-context examples to guide LLMs toward accurate responses. While these methods demonstrate the benefits of language-augmented vision, they are not tailored for vocabulary-free fine-grained recognition. 


\subsection{Enhanced Textual Embeddings for VLMs}



The performance of vision-language models (VLMs), such as CLIP \cite{CLIP}, in open-set and fine-grained recognition tasks is highly dependent on the quality of the textual embeddings used for classification. Prior work has explored several strategies for providing effective textual inputs to these models.
One common line of work involves trainable approaches, such as prompt tuning or adapter-based methods~\cite{zhou2022large, gao2024clip,10205226}, where a set of learnable textual prompts is optimized using annotated training data. These methods typically operate in a few-shot setting, requiring supervision in the form of labeled images. While effective, they rely on the availability of high-quality annotations and become unreliable in vocabulary-free settings where class names or labels are not available.
In contrast, manual prompt engineering remains a widely adopted approach for zero-shot use of CLIP, where simple templates such as “a photo of a \{class\}” are used to construct class embeddings. However, these handcrafted templates may fail to capture the fine-grained distinctions required for nuanced tasks, especially when classes are highly similar or the dataset is domain-specific.
To address the limitations of manual design, more recent approaches have explored prompt generation using large language models ~\cite{menon2022visual, cupl}, where contextual descriptions or captions are created automatically given known class names or metadata. While this improves the expressiveness of the textual prompts, these methods often assume the availability of class names, label taxonomies, or structured prior knowledge, which is an assumption that breaks down in truly vocabulary-free scenarios.
Finally, some methods attempt to expand or refine the vocabulary space by retrieving descriptions from external sources such as Wikipedia or WordNet~\cite{WordNet}, yet these too rely on known class identities and often require additional retrieval or filtering steps.
In contrast to all the above, our approach does not assume access to class names or labelled data. Instead, we generate both candidate class names and rich contextual descriptions automatically from unlabelled images using a VQA-LLM pipeline. This enables a fully vocabulary-free and training-free setup that produces more semantically grounded textual embeddings without human intervention.
\section{Method}

\begin{figure*}[!t]
  \centering
  \includegraphics[width=0.95\textwidth]{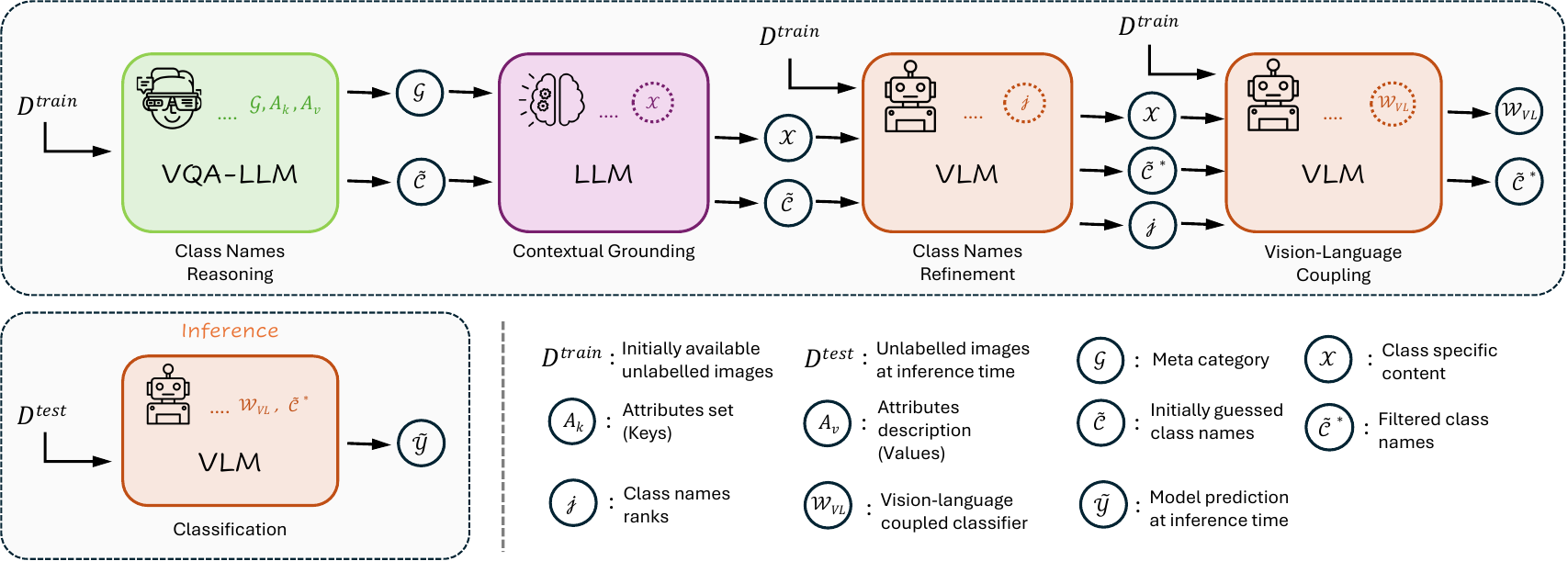}
  \vspace{-5.0pt}
  \caption{Architecture of our \our, utilized for vocabulary-free fine-grained image recognition. 
  The pipeline begins with an unlabelled training set \( \mathcal{D}_{\text{train}} \), containing a small number of unannotated images per class. 
  Next, using this semantic context, an LLM performs class name reasoning to generate an initial set of candidate class names \( \tilde{\mathcal{C}} \). Additionally, for each candidate name, the LLM also produces a set of class-specific contextual descriptions \( \mathcal{X} \) to enrich the semantic grounding of the class names. These candidates are then ranked and filtered by a class name refinement module based on their alignment with images in \( \mathcal{D}_{\text{train}} \).
  With these filtered names \( \tilde{\mathcal{C}^*} \) and contextual prompts, we construct a vision-language coupled classifier \( \mathcal{W}_{\text{VL}} \), dynamically assembled in a zero-shot manner at inference.
  }
  \label{fig:main_architecture}
  \vspace{-10.0pt}
\end{figure*}

Our method, \our{}, is designed for fully automated, training-free, vocabulary-free fine-grained image recognition. Given a small, unlabeled dataset \( \mathcal{D}_{\text{train}} \), our pipeline generates semantic class names and associates them with fine-grained visual categories, enabling recognition on an unseen test set \( \mathcal{D}_{\text{test}} \). Figure \ref{fig:main_architecture} provides an overview of the approach, and in the following, we describe each component in detail.


\subsection{Problem Formulation}

We address the task of fine-grained visual recognition (FGVR) in a vocabulary-free setting. Given a training dataset \( \mathcal{D}_{\text{train}} \) consisting of a small number of unlabeled images of fine-grained object categories, our goal is to predict semantic class names for images in a separate test set \( \mathcal{D}_{\text{test}} \). Unlike traditional FGVR setups,  
we do not assume access to any ground-truth labels or expert-provided auxiliary annotations during training  \cite{choudhury2023investigating}. Each category in \( \mathcal{D}_{\text{train}} \) is represented by only a few unlabelled samples, making this setting inherently low-resource and challenging  \cite{li2023emergent}.
Formally, given a test image \( x \in \mathcal{D}_{\text{test}} \), the objective is to assign a semantic class label \( c \in \mathcal{C} \), where \( \mathcal{C} \) is the set of fine-grained category names that are not known a priori. This vocabulary-free condition distinguishes our setting from conventional zero-shot learning, where the label set is predefined at inference time. The set \( \mathcal{C} \) must be discovered automatically based 
on the information available in \( \mathcal{D}_{\text{train}} \).
%
%

\subsection{Class Names Reasoning}

Following the common practice \cite{finer, waffle}
, we first extract potentially useful visual information from the existing unlabelled samples using a general-purpose, human-free pipeline.
To extract coarse-grained semantic context, we employ a Visual Question Answering model (VQA) integrated with a Large Language Model (LLM). The goal is twofold: (1) identify a high-level meta-category \( \mathcal{G} \) that captures the dataset’s general domain (e.g., birds, flowers), and (2) extract a structured set of fine-grained visual attributes in the form of key-value pairs \( (\mathcal{A}_k, \mathcal{A}_v) \). These attributes help the system understand which visual characteristics are relevant for distinguishing between subordinate-level categories in the dataset.
Using the meta-category \( \mathcal{G} \) and the extracted attributes, an LLM is prompted to produce an initial set of candidate class names \( \tilde{\mathcal{C}} \). 

\subsection{Class-specific Contextual Grounding}


In addition to class names, we prompt the LLM to generate a corresponding set of class-specific in-context descriptions \( \mathcal{X} \), such as defining characteristics or contextual cues (e.g., “a small songbird with yellow breast and dark wings” for Blue-winged Warbler). These descriptions later serve as rich textual prompts for vision-language matching.




\subsubsection{Generating Customized Prompts}
Here, we aim to use the LLM's knowledge gained from a vast corpus of diverse text data to generate unique sentences with a class name used in class-specific contexts. 
Specifically, these generated sentences $\mathcal{X}$ are tailored to include specific class names and their relevant descriptors, ensuring that they are closely aligned with the target categories.
Another key aspect of our approach is the generation of multiple ($M$) in-context descriptions for each class, which ensures a comprehensive coverage of linguistic variations and nuances associated with the specific class. This diversity in descriptions is crucial for capturing the full spectrum of features that define a class, enhancing the robustness of the classification algorithm \cite{cupl}.
Unlike previous approaches \cite{waffle, cupl}, our sentence generation process is fully automated, dataset-independent, and does not require domain knowledge or any human intervention, which makes it highly suitable for the vocabulary-free setup.
%
The details of LLM prompting are specified in Appendix \ref{sec:app.prompt}.

\subsubsection{Utilizing Customized Prompts}


Each of these in-context sentences, already including a given category name $c \in \tilde{\mathcal{C}}$, is embedded via the text encoder $f_T(\cdot)$. All these sentences completed with the same category name are averaged and normalised to obtain a contextual text embedding 
$\mathbf{t}_c$ as:
\vspace{-7.0pt}
\begin{equation}
    \mathbf{t}_c = \frac{1}{M} \sum_{i=1}^{M} \frac{f_T(x_i^c)}{\|f_T(x_i^c)\|},
\end{equation}
\vspace{-10.0pt}

\noindent where $x_i^c \in \mathcal{X}$ is the $i$-th in-context sentence for class $c$, $M$ is the total number of generated contextual sentences (e.g., 100), $\|\cdot\|$ denotes the $\ell_2$-norm.

In this way, to classify the images in a dataset, for each predicted class name, we obtain a single ensembled text embedding by averaging embeddings from the text CLIP encoder for all class-specific sentences generated by the LLM.

\subsection{Advanced Class Names Refinement}



Although the LLM generates plausible candidate class names in \( \tilde{\mathcal{C}} \), not all of them are visually relevant to the dataset. To address this, we use a vision-language model, such as CLIP, to compare each class-specific context \( x \in \mathcal{X} \) with visual features extracted from \( \mathcal{D}_{\text{train}} \). 
For each class name $c \in \tilde{\mathcal{C}}$, we compute its visual relevance score based on cosine similarity with each image embedding $\mathbf{v}_j \in D_{\text{train}}$:
\vspace{-5.0pt}
\begin{equation}
    \text{score}(c) = \frac{1}{N} \sum_{j=1}^{N} \cos(\mathbf{t}_c, \mathbf{v}_j) = \frac{1}{N} \sum_{j=1}^{N} \frac{\mathbf{t}_c^\top \mathbf{v}_j}{\|\mathbf{t}_c\| \cdot \|\mathbf{v}_j\|}
\end{equation}
\vspace{-10.0pt}

\noindent This process helps bridge the semantic gap between raw textual labels and the visual modality.

Lastly, from the ranked list of candidate names, we apply a soft filtering strategy to obtain a refined set of classes $\tilde{\mathcal{C}^*} \subseteq \tilde{\mathcal{C}}$ by retaining the ones with top-k scores.
Unlike traditional methods that rely on top-1 predictions or hard thresholds, our approach retains multiple high-confidence candidates that may describe different aspects or variants of the same visual concept. This filtering process improves recall, reduces semantic over-pruning, and preserves label diversity, therefore making the final classifier more robust, especially in fine-grained settings with visual ambiguities.
By doing this, unlike similar methods, we allow more class names to be used at the final classification step, which in turn allows more room for improvement at the next stages.


\subsection{Vision-Language Prompt Coupling}


In order to further increase the similarity score for potential images of the same class, we carry out coupling of text and visual embeddings. An important note is that only images with automatically assigned labels are used for this, so that no manual involvement is required at this stage.
%
%
%
Using the refined class names in \( \tilde{C^*} \), a text-based classifier is built for each class \( c \in \tilde{C^*} \) via the text encoder of the vision-language model (VLM). To address potential ambiguities in class names, as noted by \cite{finer}, we also incorporate vision embeddings. Specifically, images in the training set \( D_{train} \) are pseudo-labelled by assigning them to the class in \( \tilde{C^*} \) with the highest cosine similarity, resulting in a set of \( U_c \) pseudo-labelled samples per class. Since \( D_{train} \) contains only a few images per class, this can lead to biased visual features and limited colour or perspective variations for a specific car model. To mitigate this, we apply a random data augmentation pipeline \( K \) times per image, and further use the augmented features to construct the visual classifier. We set \( K = 10 \), following \cite{finer}.
In this way, each visual classifier prototype $\mathbf{v}_c$ for class $c \in \tilde{\mathcal{C}^*}$ is computed by averaging the augmented embeddings:
\vspace{-10.0pt}
\begin{equation}
    \mathbf{v}_c = \frac{1}{K \cdot |\mathcal{U}_c|} \sum_{i=1}^{|\mathcal{U}_c|} \sum_{k=1}^{K} \frac{f_V(\text{Aug}_k(\mathbf{x}_i^c))}{\|f_V(\text{Aug}_k(\mathbf{x}_i^c))\|},
\end{equation}
\vspace{-8.0pt}

\noindent where $\mathcal{U}_c$ is the set of pseudo-labelled images assigned to class $c$, $\text{Aug}_k$ is the $k$-th augmentation function (e.g., random crop or flip), $f_V$ is the visual encoder from the VLM.
Different to other approaches, our augmentations list includes only random crop and random flipping, which is similar to the augmentations used in training the CLIP models \cite{CLIP}. This makes resulting image embeddings more similar to the vision embeddings at inference time. 

Finally, we combine both the text- and vision-based classifiers into a coupled vision-language classifier ($W_{VL}$) for each class $c$, allowing the model to benefit from complementary information across modalities:
\vspace{-5.0pt}
\begin{equation}
    W_{VL}^{(c)} = \alpha \cdot \mathbf{t}_c + (1 - \alpha) \cdot \mathbf{v}_c ,
\end{equation}
\vspace{-15.0pt}

\noindent where $\alpha \in [0,1]$ is a weighting hyper-parameter fixed as $\alpha = 0.7$ in our experiments.


\subsection{Inference}


At inference time, given a test image \( x \in \mathcal{D}_{\text{test}} \), we compute its visual embedding and use \( \mathcal{W}_{\text{VL}} \) to predict a label from the refined class name set \( \tilde{\mathcal{C}}^* \):
\vspace{-8.0pt}
\begin{equation}
    \tilde{y} = \arg\max_{c \in \tilde{\mathcal{C}^*}} \cos\left( f_V(\mathbf{x}), W_{VL}^{(c)} \right) ,
\end{equation}
\vspace{-13.0pt}

\noindent where cosine similarity is again used to measure alignment between the test image and each coupled class representation.
Notably, the predicted output \( \tilde{y} \) is a semantic class name, not just an index, making the system interpretable and suitable for real-world applications. Crucially, the entire process, from class name discovery to inference, is fully automated, training-free, and vocabulary-free, requiring no human intervention or predefined class list.

%

\begin{table*}[!ht]
    \centering
    \normalsize
    \setlength{\tabcolsep}{4pt}
    \renewcommand{\arraystretch}{1.1}
    \begin{adjustbox}{max width=\textwidth}
    \begin{tabular}{l|cc|cc|cc|cc|cc|cc}
    \toprule
    \textbf{} & \multicolumn{2}{c|}{\textbf{Birds-200}} & \multicolumn{2}{c|}{\textbf{Cars-196}} & \multicolumn{2}{c|}{\textbf{Dogs-120}} & \multicolumn{2}{c|}{\textbf{Flowers-102}} & \multicolumn{2}{c|}{\textbf{Pets-37}} & \multicolumn{2}{c}{\textbf{Average}} \\
    \textbf{Method} & cACC & sACC & cACC & sACC & cACC & sACC & cACC & sACC & cACC & sACC & cACC & sACC \\
    \midrule
    \textcolor{gray!70}{CLIP (zero-shot, UB) \cite{CLIP}} & \textcolor{gray!70}{57.4} & \textcolor{gray!70}{80.5} & \textcolor{gray!70}{63.1} & \textcolor{gray!70}{66.3} & \textcolor{gray!70}{56.9} & \textcolor{gray!70}{75.5} & \textcolor{gray!70}{69.7} & \textcolor{gray!70}{77.8} & \textcolor{gray!70}{81.7} & \textcolor{gray!70}{87.8} & \textcolor{gray!70}{65.8} & \textcolor{gray!70}{77.6} \\
    \midrule
    CLIP-Sinkhorn \cite{chen2022plot} & 23.5 & - & 18.1 & - & 12.6 & - & 30.9 & - & 23.1 & - & 21.6 & - \\
    DINO-Sinkhorn \cite{chen2022plot} & 13.5 & - & 7.4 & - & 11.2 & - & 17.9 & - & 5.2 & - & 19.1 & - \\
    KMeans \cite{krishna1999genetic} & 36.6 & - & 30.6 & - & 16.4 & - & 66.9 & - & 32.8 & - & 36.7 & - \\
    WordNet \cite{WordNet} & 39.3 & 57.7 & 18.3 & 33.3 & 53.9 & \textbf{70.6} & 42.1 & 49.8 & 55.4 & 61.9 & 41.8 & 54.7 \\
    BLIP-2 \cite{blip-2} & 30.9 & 56.8 & 43.1 & 57.9 & 39.0 & 58.6 & 61.9 & \textbf{59.1} & 61.3 & 60.5 & 47.2 & 58.6 \\
    CLEVER \cite{DBLP:journals/corr/abs-2111-03651} & 7.9 & - & - & - & - & - & 6.2 & - & - & - & - & - \\
    SCD \cite{han2024whatsnameclassindices} & 46.5 & - & - & - & \textbf{57.9} & - & - & - & - & - & - & - \\
    CaSED \cite{conti2024vocabularyfreeimageclassification} & 25.6 & 50.1 & 26.9 & 41.4 & 38.0 & 55.9 & \textbf{67.2} & 52.3 & 60.9 & 63.6 & 43.7 & 52.6 \\
    FineR \cite{finer} & \underline{51.1} & \underline{69.5} & \underline{49.2} & \underline{63.5} & 48.1 & 64.9 & 63.8 & 51.3 & \textbf{72.9} & \underline{72.4} & 57.0 & 64.3 \\
    \rowcolor{cyan!10}
    \midrule
    \textit{\our{} (Ours)} & \textbf{52.1} & \textbf{70.1} & \textbf{51.2} & \textbf{64.0} & \underline{51.8} & \underline{67.1} & \underline{64.8} & \underline{54.0} & \underline{71.7} & \textbf{76.2} & \textbf{58.4} & \textbf{66.3} \\
    %
    $\Delta$ to baseline, absolute \% & \cellcolor{green!15}+1.0 \% & \cellcolor{green!15}+0.6 \% & 
    \cellcolor{green!15}+2.0 \% & \cellcolor{green!15}+0.5 \% & \cellcolor{green!15}+3.7 \% & \cellcolor{green!15}+2.2 \% & \cellcolor{green!15}+1.0 \% & \cellcolor{green!15}+2.7 \% & 
    \cellcolor{green!05}-1.2 \% &
    \cellcolor{green!15}+3.8 \% & \cellcolor{green!15}+1.4 \% & \cellcolor{green!15}+2.0 \% \\
    $\Delta$ to baseline, relative \% & \cellcolor{green!15}+2.0 \% & \cellcolor{green!15}+0.9 \% & 
    \cellcolor{green!15}+4.1 \% & \cellcolor{green!15}+0.8 \% & \cellcolor{green!15}+7.7 \% & \cellcolor{green!15}+3.4 \% & \cellcolor{green!15}+1.6 \% & \cellcolor{green!15}+5.3 \% & 
    \cellcolor{green!05}-1.6 \% &
    \cellcolor{green!15}+5.3 \% & \cellcolor{green!15}+2.5 \% & \cellcolor{green!15}+3.2 \% \\
    \bottomrule
    \end{tabular}
    \end{adjustbox}
    \vspace{-0.1in}
    \caption{Vocabulary-free classification performance comparison on fine-grained datasets using cACC (clustering accuracy, \%) and sACC (semantic accuracy, \%). For our approach and FineR, the class discovering is done with 3 unlabelled images per class, and the results are averaged across 10 runs. CLIP zero-shot results are with known 
    class names. 
    Best results are in bold, second-best are underlined.}
    \label{tab:fgvr-results}
    \vspace{-13.0pt}
\end{table*}

\section{Experiments and Analysis}

\subsection{Experimental setup}

\paragraph{Datasets.}
%
We evaluate our method under the vocabulary-free setting across five standard fine-grained visual recognition (FGVR) benchmarks: CUB-200 (Birds-200) \cite{wah2011caltech}, Stanford Cars (Cars-196) \cite{hu2025car}, Stanford Dogs (Dogs-120) \cite{khosla2011novel}, Oxford Flowers (Flowers-102) \cite{nilsback2008automated}, and Oxford Pets (Pets-37) \cite{patino2016pets}.
During the evaluation, we follow the vocabulary-free setup proposed in \cite{finer}.
Specifically, in our default setting, a low-resource scenario is simulated by limiting the number of unlabelled images per category in the training set \( D_{\text{train}} \) to just 3, randomly sampled from each dataset’s training split. These samples are used for class name discovery. The test split of each dataset serves as the evaluation set \( D_{\text{test}} \) with no overlap between the training and test sets, i.e., \( D_{\text{train}} \cap D_{\text{test}} = \emptyset \).
\vspace{-10.0pt}

\paragraph{Evaluation Metrics.}
%
Due to the unsupervised nature of the vocabulary-free task, an exact one-to-one alignment between the predicted class names in \( \tilde{C^*} \) and the ground truth categories in \( C \) cannot be assumed \cite{CLIP, finer}. To evaluate performance effectively, we use two complementary metrics: Clustering Accuracy (cACC) and Semantic Accuracy (sACC). cACC measures how well the model groups images from the same category, regardless of the semantic correctness of the assigned labels. To address this limitation, sACC evaluates the semantic similarity between predicted class names and the true categories using a pre-trained language model. This allows for a more nuanced assessment of classification quality, capturing how close or far off a prediction is in meaning. For instance, assigning a closely related term would be penalized less than a semantically distant mislabel. Together, cACC and sACC provide a holistic view of performance by ensuring that clusters are not only visually coherent but also semantically meaningful.
\vspace{-10.0pt}
\paragraph{Baselines and Implementation Details.}
%
For baseline details and implementation setup, refer to App. \ref{sec:app.baselines} - \ref{sec:app.implement}.

\subsection{Quantitative analysis}

\subsubsection{Vocabulary-free Setting}

In this setting, 
the performance is reported using both clustering accuracy (cACC) and semantic accuracy (sACC) metrics. Our approach is compared with a suite of competitive baselines, including unsupervised clustering (KMeans, Sinkhorn variants\cite{chen2022plot}), retrieval-based classification (WordNet \cite{WordNet}, CaSED \cite{conti2024vocabularyfreeimageclassification}), vision-language QA (BLIP-2)  \cite{blip-2}, and current SOTA open-vocabulary method FineR  \cite{finer}.

As summarized in Table \ref{tab:fgvr-results}, our method achieves the highest overall performance with an average of 58.4\% cACC and 66.3\% sACC on all five datasets. Notably, \our{} approach, on average, surpasses FineR (which is also our baseline) by +2.5\% and +2.0\%, respectively. It obtains the best results in 4 out of 5 datasets on cACC, and 3 out of 5 datasets on sACC. Our model is particularly strong in datasets with subtle visual differences but semantically rich categories, such as Dogs-120 (+7.7\% cACC), Cars-196 (+4.1\%), and Flowers-102 (+5.3\% sACC). These gains highlight our model’s ability to leverage latent visual and linguistic cues without relying on predefined vocabularies.

Despite the overall strong performance, there are cases where alternative approaches outperform our method. For example, WordNet achieves the highest semantic accuracy on Dogs-120 (70.6\%), slightly above ours (67.1\%). This may be attributed to WordNet’s rich lexical coverage and precise taxonomy for dog breeds, which can align well with the visually grounded CLIP embeddings when domain-specific class names are present.
BLIP-2 outperforms our approach on Flowers-102 in cACC (66.9\% vs. 64.8\%), suggesting that direct VQA-based object naming might be more effective in domains with well-photographed, object-centric imagery and relatively simple visual context.
On Pets-37, FineR\cite{finer}, which is our baseline, yields higher clustering accuracy (72.9\% vs. 71.7\%). However, we still outperform it on sACC (76.2\% vs. 72.4\%), indicating that our clusters, while slightly less tight, are semantically more coherent.
These findings suggest that while specialized retrieval or QA pipelines can sometimes exploit structured vocabularies more effectively in narrow domains, our method offers a more robust and generalizable framework across diverse datasets and domain types. The combination of high cACC and sACC underscores its strength in both grouping visually similar instances and naming them accurately, even in challenging fine-grained scenarios.

Additional ablation study for the Class-specific Contextual Grounding (CCG) and Class Names Refinement (CNR) components of our \our{} can be found in Appendix \ref{sec:app.ablation}.

\subsubsection{Zero-shot Setting}

In addition to the targeted vocabulary-free setup, we discover that our approach can be successfully transferred to zero-shot scenarios.
In Table \ref{tab:0shot-results}, we evaluate \our{} under the zero-shot classification setting across four datasets: Stanford Cars (Cars), Oxford Flowers (Flowers), and Oxford Pets (Pets) along with benchmarking it on the ImageNet dataset. The comparison includes two method categories: (i) Train-free, which does not involve any training, and (ii) Automated, which requires no human intervention at any stage. Our method falls into both these categories, reflecting its practical usability and scalability.
Among all automated and train-free baselines, \our{} consistently achieves competitive performance, obtaining the best or second-best accuracy across all datasets. Notably, \our{} secures the highest accuracy among automated methods on Flowers (72.04\%) and Pets (90.30\%), while achieving 68.68\% on ImageNet and 63.92\% on Cars. These results closely rival the best-performing training-based methods, such as ProText and CuPL, which use additional supervision or prompt tuning. For example, ProText marginally outperforms \our{} on Flowers (74.42\%) and Pets (92.72\%), but at the cost of requiring manual effort or training, thus making them less scalable for real-world deployment.
Furthermore, within the automated + train-free category, \our{} surpasses all baselines, including CLIP (default), DCLIP, and WaffleCLIP, across three out of four datasets. For instance, it improves over CLIP (default) by +1.36\% on ImageNet, +0.17\% on Flowers, and +2.25\% on Pets, highlighting the effectiveness of our refined class discovery and naming approach even in the absence of training or human-authored prompts.
\vspace{-5.0pt}

\begin{table}[!h]
    \centering
    \normalsize
    \setlength{\tabcolsep}{4pt}
    \renewcommand{\arraystretch}{1.1}
    \begin{adjustbox}{max width=\textwidth}
    \begin{tabular}{l|c|c|c|c|c|c}
    \toprule
    \small{\textbf{Method}} & \small{Train-} & \small{Auto-} & \small{Image-} & \small{Cars} & \small{Flow-} & \small{Pets}  \\
     & \small{free} & \small{mated} & \small{Net} &  & \small{ers} &   \\ 
     
    \midrule
    \scriptsize{CLIP} \cite{CLIP} & \small{\cellcolor{green!15}\checkmark} & \small{\cellcolor{red!10}$\times$} & \small{68.35} & \small{64.76} & \small{70.40} & \small{89.13} \\
    \scriptsize{ProText} \cite{khattak2024learning} & \small{\cellcolor{red!10}$\times$} & \small{\cellcolor{red!10}$\times$} & \small{\textbf{69.80}} & \small{\textbf{66.77}} & \small{\textbf{74.42}} & \small{\textbf{92.72}} \\
    \scriptsize{CuPL} \cite{cupl} & \small{\cellcolor{green!15}\checkmark} & \small{\cellcolor{red!10}$\times$} & \small{69.62} & \small{65.95} & \small{73.85} & \small{91.11} \\
    \midrule
    \scriptsize{CLIP} \scriptsize{(default)} & \small{\cellcolor{green!15}\checkmark} & \small{\cellcolor{green!15}\checkmark} & \small{66.72} & \small{63.75} & \small{67.34} & \small{88.25} \\
    \scriptsize{DCLIP} \cite{menon2022visual} & \small{\cellcolor{green!15}\checkmark} & \small{\cellcolor{green!15}\checkmark} & \small{68.03} & - & - & \small{86.92} \\
    \scriptsize{WaffleCLIP} \cite{waffle} & \small{\cellcolor{green!15}\checkmark} &  \small{\cellcolor{green!15}\checkmark} & \small{68.34} & \small{63.42} & \small{72.00} & \small{89.57} \\
    \rowcolor{cyan!10}    
    \scriptsize{\textit{\our{}} \scriptsize{(Ours)}} & \small{\cellcolor{green!15}\checkmark} & \small{\cellcolor{green!15}\checkmark} & \small{\textbf{68.68}} & \small{\textbf{63.92}} & \small{\textbf{72.04}} & \small{\textbf{90.30}}  \\
    \bottomrule
    \end{tabular}
    \end{adjustbox}
    \vspace{-5.0pt}
    \caption{Zero-shot classification performance comparison on fine-grained datasets and ImageNet. 
    The categories include "Train-free" (no extraining is required) and "Automated" (no human intervention is required at any stage). 
    "CLIP (default)" is not using any manual prompts. 
    Best results are in bold for each group.}
    \label{tab:0shot-results}
    \vspace{-8.0pt}
\end{table}

These findings underscore \our{}’s capability to serve as a strong and practical zero-shot learner, balancing accuracy with scalability and ease of use, essential qualities for deployment in the vocabulary-free setting.

\begin{figure*}[!t]
  \centering
  \includegraphics[width=0.95\textwidth]{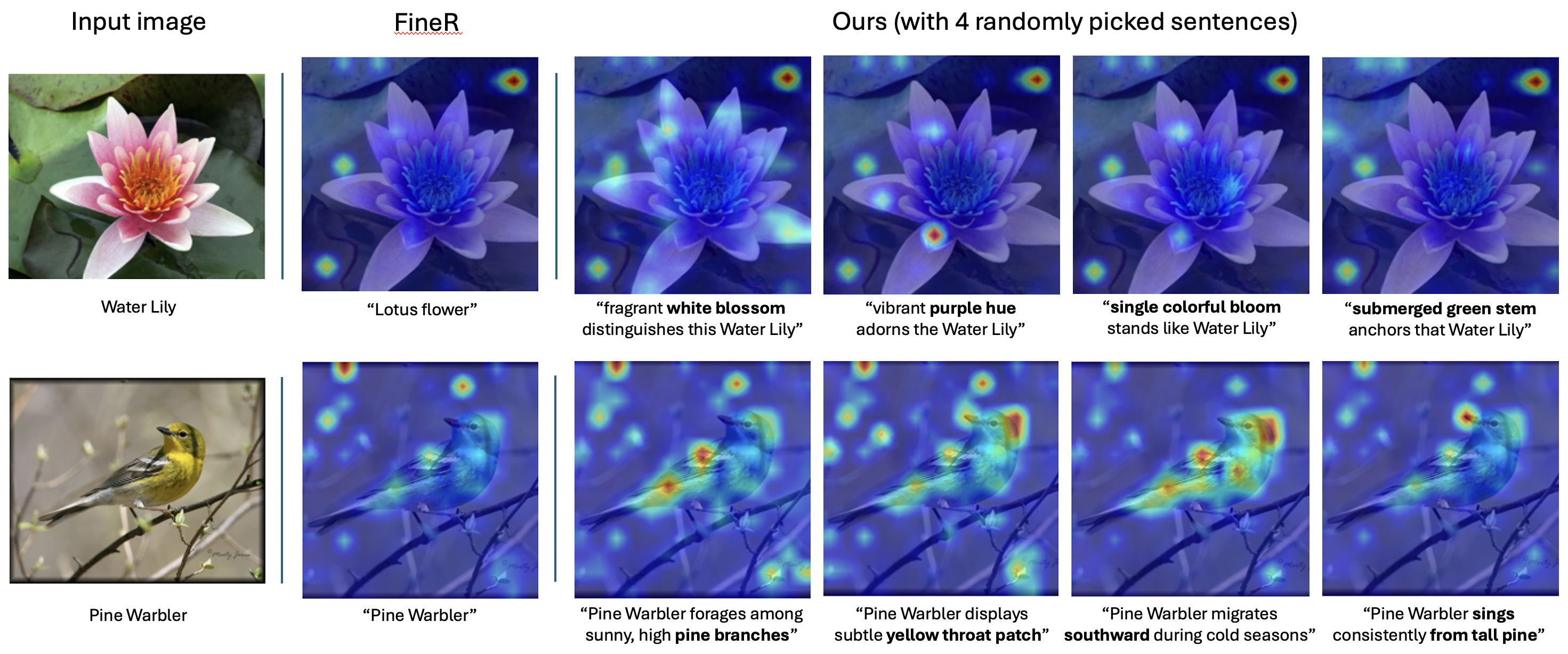}
  \vspace{-6.0pt}
  \caption{Qualitative comparison of attention heat maps generated by previous state-of-the-art method and our \our{} on samples from different fine-grained datasets.
  For our approach, we randomly pick 4 generated class-specific contexts, while for the competitor, the only utilized context is the class name itself.
  Visually helpful words matching the actual visual scene in the input image are highlighted in bold.
  }
  \vspace{-12.0pt}
  \label{fig:qual_heat-maps}
\end{figure*}

\subsubsection{Few-shot Setting}
Additionally, we verify the successful applicability of our approach to the few-shot scenarios.
Table \ref{tab:fewshot-results} presents a comparison of few-shot classification performance across three datasets: ImageNet, Cars, and Pets. Each model is evaluated using 16 images per class for ImageNet and 8 per class for the fine-grained datasets. Similar to earlier evaluations, methods are grouped based on whether they are Train-free (i.e., no learning stage) and Automated (i.e., no manual intervention).
To our knowledge, there is no train-free and fully automated approach for few-shot classification.

\begin{table}[!h]
    \centering
    \normalsize
    \setlength{\tabcolsep}{4pt}
    \renewcommand{\arraystretch}{1.1}
    \begin{adjustbox}{max width=\textwidth}
    \begin{tabular}{l|c|c|c|c|c}
    \toprule
    \textbf{Method} & Train- & Auto- & Image- & Cars & Pets  \\
     & free & mated & Net &  &   \\    
    \midrule
    \small{CLIP} \scriptsize{(linear probe)} & \cellcolor{red!10}$\times$ & \cellcolor{green!15}\checkmark & 67.31 & 73.67 & 85.34 \\
    \small{CLIP} \scriptsize{+ Ours} & \cellcolor{green!15}\checkmark & \cellcolor{red!10}$\times$ & 70.12 & - & - \\    
    \small{CoOp} \cite{Zhou_2022} & \cellcolor{red!10}$\times$ & \cellcolor{green!15}\checkmark & 71.87 & 79.30 & 91.87 \\
    \small{CoCoOp} \cite{zhou2022conditionalpromptlearningvisionlanguage} & \cellcolor{red!10}$\times$ & \cellcolor{green!15}\checkmark & 70.83 & 70.44 & 93.34 \\
    \small{PromptSRC} \cite{khattak2023selfregulatingpromptsfoundationalmodel} & \cellcolor{red!10}$\times$ & \cellcolor{green!15}\checkmark & \textbf{73.17} & \textbf{80.97} & \textbf{93.67} \\    
    \small{MaPLe} \cite{khattak2023maplemultimodalpromptlearning} & \cellcolor{red!10}$\times$ & \cellcolor{green!15}\checkmark & 72.33 & 79.47 & 92.83 \\
    \midrule
    \small{CLIP} \scriptsize{(default) + Ours} & \cellcolor{green!15}\checkmark & \cellcolor{green!15}\checkmark & 69.58 & - & - \\
    \rowcolor{cyan!10}    
    \small{\textit{\our{}} \scriptsize{(Ours)}} & \cellcolor{green!15}\checkmark & \cellcolor{green!15}\checkmark & \textbf{70.59} & \textbf{72.24} & \textbf{92.53}  \\
    \bottomrule
    \end{tabular}
    \end{adjustbox}
    \vspace{-5.0pt}
    \caption{Few-shot classification performance comparison on fine-grained datasets and ImageNet. 
    The categories include "Train-free" (no training is required) and "Automated" (no human intervention is required at any stage). 
    "CLIP (default)" is not using any manual prompts. 
    Best results are in bold for each group.}    
    \label{tab:fewshot-results}
    \vspace{-15.0pt}
\end{table}

Our method, \our{}, achieves strong results despite being both train-free and fully automated. Specifically, it outperforms all other train-free baselines across all datasets: 70.59\% on ImageNet, 72.24\% on Cars, and 92.53\% on Pets. Notably, \our{} surpasses the popular CLIP (default) baseline by +1.01\% on ImageNet and +1.25\% on Pets, demonstrating the benefit of leveraging refined class descriptions even in low-data regimes.
Although training-based methods like PromptSRC \cite{khattak2023selfregulatingpromptsfoundationalmodel} and MaPLe \cite{khattak2023maplemultimodalpromptlearning} slightly edge out \our{} on ImageNet (by ~2.5\%) and Cars (by ~8.7\%), they require prompt tuning or additional supervision, therefore making them less scalable and more resource-intensive. Similarly, CoOp and CoCoOp, which involve training class-specific prompts, outperform \our{} on Cars but still fall short on Pets.
An interesting observation is that CLIP + Ours (a simple augmentation of CLIP with our method) already shows improved performance over the CLIP linear probe baseline on ImageNet (+2.81\%), suggesting that our class refinement pipeline is complementary to existing approaches.

These results reinforce the utility of \our{} as a highly effective and scalable few-shot learner, offering strong performance while maintaining zero manual supervision or training. It makes our approach suitable for real-world deployment in dynamic or under-resourced settings.

\subsection{Qualitative Analysis}


\subsubsection{Predictions Analysis}

Figure \ref{fig:qual_predictions} illustrates qualitative comparisons between our method and FineR across three datasets: Oxford Flowers, CUB-200, and Stanford Dogs. These examples highlight the effectiveness of our approach in generating accurate and semantically appropriate class names in the absence of a predefined vocabulary.
In the example from Oxford Flowers, FineR predicts ``White-Centered Pink Flower", a visually descriptive label that captures superficial color and structure but lacks alignment with the dataset’s taxonomy. In contrast, our method predicts ``Purple Columbine", the correct class name. This improvement can be attributed to the integration of Class-specific Contextual Grounding, which enriches the semantic cues used during naming and helps steer the model toward more taxonomically valid concepts.
The example from Stanford Dogs dataset further illustrates this point: while FineR predicts ``Husky", our method refines this to the more specific ``Siberian Husky", aligning with the ground truth. This may be attributed to the benefit of our class name filtration mechanism, which reduces semantic ambiguity by filtering out less relevant or overly generic candidates from the label space.
In the example taken from CUB-200 dataset, FineR misclassifies a ``Black Tern" as ``Bonaparte’s Gull", both visually similar birds but distinct classes. Our method produces the correct label, which reflects the combined impact of using class-specific visual cues and filtering mechanisms to guide the naming process toward higher-granularity predictions.

Overall, these results underscore the effectiveness of our 
core components (contextual class conditioning and refined label selection) in enabling precise, semantically grounded predictions under vocabulary-free constraints.


\subsubsection{Attention Maps Analysis}

We present a qualitative comparison of attention heat maps between FineR (baseline) and our proposed method, \our{}, on samples from two fine-grained datasets (see Figure \ref{fig:qual_heat-maps}). The figure illustrates how each method attends to image regions when conditioned on different textual inputs. FineR relies solely on the class name, while our method utilizes class-specific context sentences generated automatically, four of which are randomly selected for visualization.

In the first example (Water Lily), FineR’s attention is diffuse and partially aligned, focusing on the flower's general structure. However, our method,conditioned on richer contexts like “fragrant white blossom” and “submerged green stem anchors”, produces heat maps that better align with semantically relevant visual features, such as the flower center, stem, and bloom shape. The class-specific textual prompts help ground the model’s attention in parts of the image that distinguish the class more effectively.
In the second example (Pine Warbler), FineR’s attention is again relatively coarse, whereas our model focuses on distinctive regions such as the “yellow throat patch” or the “pine branches” when prompted with corresponding textual cues. This demonstrates that incorporating rich, attribute-informed language enhances the model’s interpretability and fine-grained discrimination, guiding attention to visually meaningful regions.
These results indicate that our approach benefits not only classification performance but also visual-semantic alignment, as it better captures the relationship between fine-grained attributes and image regions through contextual grounding.

\section{Conclusion}

We proposed \our{}, an automated, training-free framework for fine-grained image recognition in the vocabulary-free setting. It addresses key limitations of prior methods by eliminating the need for predefined class vocabularies, expert annotations, or manual prompt engineering. Instead, \our{} leverages the synergy between large language models and vision-language models to perform class discovery and classification in a semantically grounded and scalable manner.
Our framework is built upon three core components: (1) Class-specific Contextual Grounding, which automatically generates rich and descriptive class-specific prompts; (2) Advanced Class Name Filtration, which preserves semantically plausible class candidates beyond top-1 predictions; and (3) an enhanced recognition pipeline that dynamically assembles vision-language classifiers without any retraining or supervision. Together, these modules enable our model to disambiguate visually similar categories, improve interpretability, and handle vocabulary-free fine-grained scenarios with minimal assumptions.
Extensive experiments demonstrate that \our{} achieves state-of-the-art performance on the vocabulary-free task and in zero-shot and few-shot classification among training-free and fully-automated methods. This offers a practical solution for real-world deployment in environments where class labels are unknown, evolving, or difficult to \nobreak{annotate}.
%
{
    \small
    \bibliographystyle{ieeenat_fullname}
    \bibliography{main}
}
\clearpage

\appendix

\section{Appendix A}

\subsection{Experimental Details}

\subsubsection{Baselines}
\label{sec:app.baselines}

As fine-grained visual recognition without expert annotations is an emerging area, there are limited existing baselines, with FineR \cite{finer} and CLEVER \cite{DBLP:journals/corr/abs-2111-03651} 
being notable exceptions. To provide a comprehensive comparison, we also include below-mentioned strong baselines. 
(i) CLIP Zero-Shot Upper Bound (UB), which uses the ground-truth class names as text prompts, reflecting expert-level knowledge and serving as an upper performance bound.
(ii) WordNet Baseline, which uses CLIP with a large vocabulary of 119,000 nouns from WordNet \cite{WordNet}. 
(iii) BLIP-2  \cite{blip-2} and Flan-T5xxl \cite{lamott2024leveraging}, a VQA-based approach that identifies the main object in an image via the prompt “What is the name of the main object in this image?”. 
(iv) SCD \cite{han2024whatsnameclassindices}, 
which first clusters images and then narrows down labels using CLIP with a combined vocabulary from WordNet and Wikipedia bird names.
(v) CaSED \cite{conti2024vocabularyfreeimageclassification}, 
which retrieves captions from a large-scale knowledge base and extracts class names by parsing and classifying nouns with CLIP. 
(vi) KMeans clustering on CLIP visual features. 
(vii) Sinkhorn-Knopp Clustering, a parametric method, 
applied with features from CLIP and DINO. All baselines are evaluated using the CLIP ViT-B/16 vision encoder. 


\subsubsection{Implementation Details}
\label{sec:app.implement}

For the Class Names Reasoning and Class Names Refinement modules, following \cite{finer}, BLIP-2 \cite{blip-2} with Flan-T5xxl  \cite{flanT5xxl} are used as the visual question answering (VQA) model, ChatGPT (gpt-3.5-turbo) accessed via the OpenAI API as the large language model (LLM), and CLIP ViT-B/16 as the vision-language model (VLM). The hyperparameters for multi-modal fusion and data augmentation are set to \( \alpha = 0.7 \) and \( K = 10 \), respectively.

For the Contextual Grounding module, we use Google Gemini 2.0 LLM accessed via the public API. Where we prompt the LLM to generate 100 in-context sentences about the guessed class name. These 100 sentences per class are then averaged and normalized.
Specifics about the prompt and other details will be publicly available in our shared repository. 
In the Class Names Refinement module, a larger CLIP model based on ViT-L/14 is used for filtration, and finally, in the Vision-Language Coupling block we utilize the smaller ViT-B/16 base model for faster inference.

\subsubsection{Prompt Design}
\label{sec:app.prompt}

The exact prompt used with Gemini-2.0 large language model to obtain in-context class-specific sentences (specifically applied for the CUB-200 dataset, "classname" word will be replaced with an actual guessed label from the classname reasoning step):

\begin{figure}
\centering
  \begin{mybox}{grey}{1}
    \scriptsize{
    \begin{verbatim}
Generate 100 short and common sentences with noun 
{classname}, a type of bird, as a main subject. 

This noun should only be used in a realistic 
and descriptive general context with various real 
and related scenarios. In the sentence, highlight 
something specific about the classname, a type 
of bird, which helps to distinct it from other
birds (it can be its color, shape, size, 
background, and so on). 

Only use the main and original sense of this noun,
no idioms. Only use visually descriptive adjectives 
or participles. Each sentence should be between 
5 to 8 words (excluding the noun). Do not use 
the possessive form. Do not add an article at 
the beginning of the sentence. Do not repeat 
the noun in the same sentence. Do not capitalize 
the first letter of the sentence unless this is 
a name. Do not add a dot at the end of sentence. 
Make sure sentences are diverse and do not repeat 
each other. 

Make sure the noun is included in each sentence. 
Make sure the sentences are between 5 to 8 words each. 

Return output in the following structure as a single 
line: ["<generated_sentence_1>", 
"<generated_sentence_2>", ..., 
"<generated_sentence_n>"] 
    \end{verbatim}
    }
    \vspace{-1.5em}
  \end{mybox}

\end{figure}

\subsection{Additional Analysis}

\subsubsection{Ablation study}
\label{sec:app.ablation}



In this section we assess the contribution of our key design components: Class-specific Contextual Grounding (CCG) and Class Names Refinement (CNR). For this, we perform an ablation study on the Stanford Dogs dataset, with results shown in Table \ref{tab:ablation}. The results demonstrate that both components contribute meaningfully to performance. Adding CCG alone already improves clustering accuracy (cACC) and semantic accuracy (sACC) compared to the baseline (no CCG or CNR), increasing cACC from 51.30\% to 51.86\%, and sACC from 65.41\% to 66.98\%. This suggests that incorporating context tailored to each class helps align the discovered clusters more effectively.

\begin{table}[!h]
    \centering
    \normalsize
    \setlength{\tabcolsep}{4pt}
    \renewcommand{\arraystretch}{1.1}
    \begin{adjustbox}{max width=\textwidth}
    \begin{tabular}{cc|cc|cc}
    \toprule
    \multicolumn{2}{c|}{\textbf{Components}} & \multicolumn{2}{c|}{\textbf{Accuracy}} & \multicolumn{2}{c}{\textbf{Sensitivity}} \\
    CCG & CNR & cACC $\uparrow$ & sACC $\uparrow$ & FN $\downarrow$ & TP $\uparrow$  \\
    \midrule
    $\times$ & $\times$ & 51.30 & 65.41 & 7 & 52 \\
    \checkmark & $\times$ & 51.86 & 66.98 & 4 & 55 \\    
    \checkmark & \checkmark & \textbf{51.99} & \textbf{67.11} & \textbf{0} & \textbf{59} \\
    \bottomrule
    \end{tabular}
    \end{adjustbox}
    \vspace{-5.0pt}
    \caption{Ablation study for our proposed components. The performance is reported for the Stanford Dogs dataset for a fixed run. Acronyms are: Class-specific Contextual Grounding (CCG), Class Names Refinement (CNR), Clustering accuracy (cACC), Semantic accuracy (sACC). The number of filtered (unused) real class names is denoted as False Negative (FN), and the number of kept (used) real class names as True Positive (TP). Best results are in bold.}    
    \label{tab:ablation}
    \vspace{-10.0pt}
\end{table}




For the sensitivity analysis, we compare each guessed class name with the ground truth labels. If a guessed name fully matches any of the actual labels, then it is chosen for analysis and disregarded otherwise. Next, the class name is considered as True Positive if it was correctly guessed and used further for the classification, and as False Negative if it was correctly guessed but was filtered out at the Class Names
Refinement stage (and was not used for the classification).
It can be observed that when both CCG and CNR are enabled, the system achieves the highest accuracy and sensitivity, with cACC of 51.99\% and sACC of 67.11\%. 
Importantly, the final configuration results in zero false negatives (FN = 0) with no real class names mistakenly filtered out while retaining all 59 ground-truth class names (TP = 59). This highlights the ability of our refinement mechanism to retain all semantically relevant classes.

In summary, both CCG and CNR contribute complementary benefits: CCG enriches semantic grounding, while CNR ensures high recall in class selection. Their combination is critical for robust and precise vocabulary-free classification in fine-grained domains.

\end{document}